\documentclass[letterpaper, 10 pt, conference]{ieeeconf}  % Comment this line out if you need a4paper

\IEEEoverridecommandlockouts                              % This command is only needed if 
\usepackage{amsmath}
\usepackage{float}
\usepackage[T1]{fontenc}
\usepackage{graphics} % for pdf, bitmapped graphics files
\usepackage{amssymb}
\usepackage{graphicx}
\usepackage{algorithm}
\usepackage{booktabs}
\usepackage[table]{xcolor} % 确保正确加载xcolor宏包和table选项
\usepackage{pifont}
\usepackage{subcaption}
\usepackage{tabularx}
\usepackage{cite}
\usepackage[noend]{algpseudocode}

\definecolor{customgreen}{RGB}{0, 153, 0} % 假设的RGB值，您需要替换为准确值

\newcommand{\cmark}{{\color{customgreen}\ding{51}}} % 使用自定义绿色的勾
\newcommand{\xmark}{{\color{red}\ding{55}}}          % 使用红色的叉

\newcolumntype{Y}{>{\centering\arraybackslash}X} % 定义新的列类型Y，基于X但居中对齐
\title{\LARGE \bf
Learning Occlusion-aware Decision-making from Agent Interaction via Active Perception
}

\author{Jie Jia$^{1}$, Yiming Shu$^{2}$, Zhongxue Gan$^{1}$, Wenchao Ding$^{1}$ % <-this % stops a space
\thanks{$^{1}$Jie Jia, Zhongxue Gan and Wenchao Ding are with the Academy for Engineering and Technology, Fudan University.}
\thanks{$^{2}$Yiming Shu is with the Department of Mechanical Engineering, the University of Hong Kong.}%
\thanks{Email: dingwenchao@fudan.edu.cn, 22210860105@m.fudan.edu.cn}
}

\begin{document}

\maketitle
\thispagestyle{empty}
\pagestyle{empty}

%%%%%%%%%%%%%%%%%%%%%%%%%%%%%%%%%%%%%%%%%%%%%%%%%%%%%%%%%%%%%%%%%%%%%%%%%%%%%%%%
\begin{abstract}

Occlusion-aware decision-making is essential in autonomous driving due to the high uncertainty of various occlusions. Recent occlusion-aware decision-making methods encounter issues such as high computational complexity, scenario scalability challenges, or reliance on limited expert data. Benefiting from automatically generating data by exploration randomization, we uncover that reinforcement learning (RL) may show promise in occlusion-aware decision-making. However, previous occlusion-aware RL faces challenges in expanding to various dynamic and static occlusion scenarios, low learning efficiency, and lack of predictive ability. To address these issues, we introduce Pad-AI, a self-reinforcing framework to learn occlusion-aware decision-making through active perception. Pad-AI utilizes vectorized representation to represent occluded environments efficiently and learns over the semantic motion primitives to focus on high-level active perception exploration. Furthermore, Pad-AI integrates prediction and RL within a unified framework to provide risk-aware learning and security guarantees. Our framework was tested in challenging scenarios under both dynamic and static occlusions and demonstrated efficient and general perception-aware exploration performance to other strong baselines in closed-loop evaluations.

\end{abstract}

%%%%%%%%%%%%%%%%%%%%%%%%%%%%%%%%%%%%%%%%%%%%%%%%%%%%%%%%%%%%%%%%%%%%%%%%%%%%%%%%
\section{Introduction}

\begin{figure}[t]
  \centering 
  \includegraphics[width=0.48\textwidth, height=0.2\textheight]{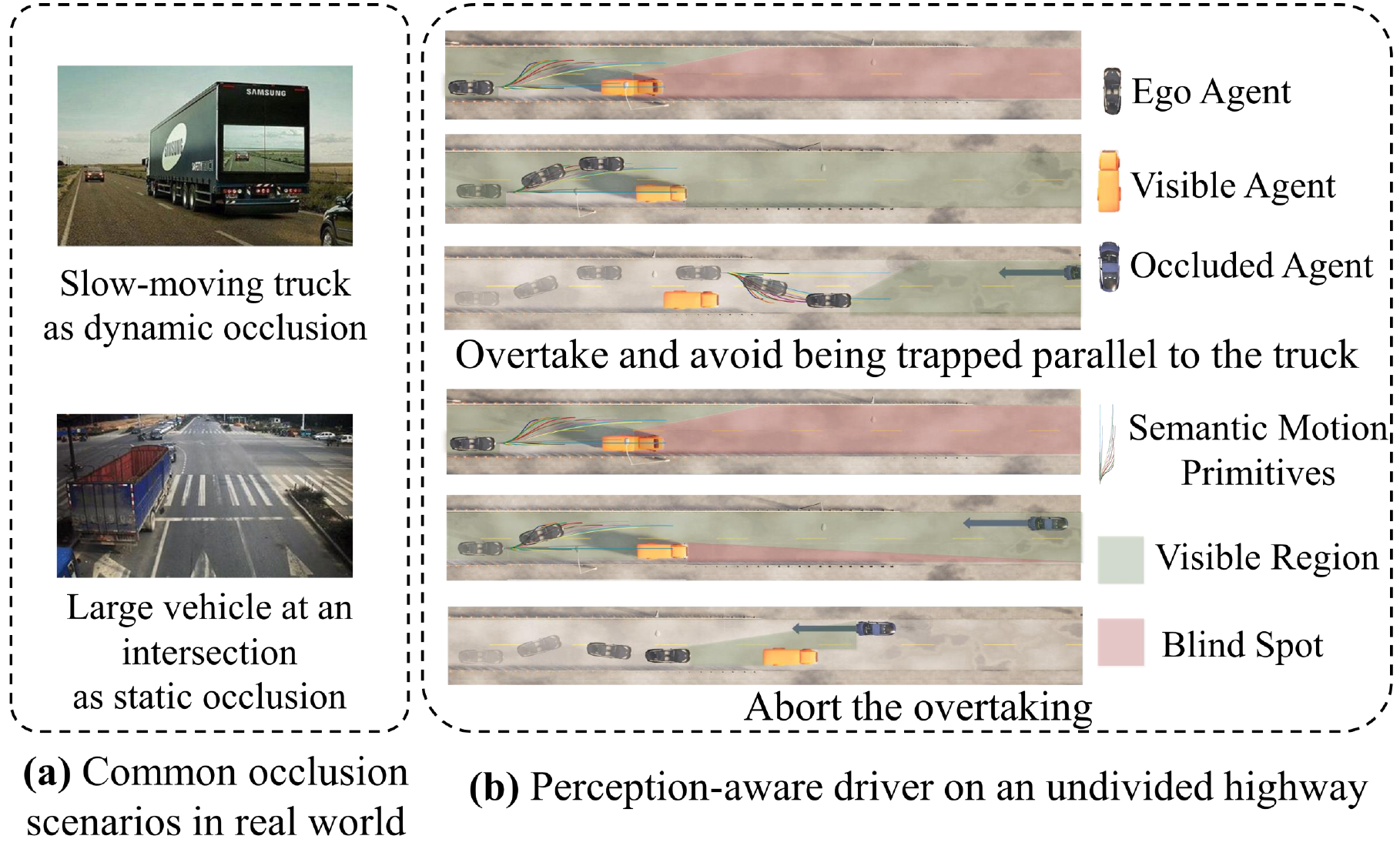}
  \caption{\textbf{(a)} Common occlusion scenarios. \textbf{(b)} Instead of conservatively following the slow truck, our method will cautiously reduce the occlusion uncertainty with an active probe motion. Based on updated observations, our method swiftly decides whether to overtake or abort the overtaking.}
  \label{cover}
  \vspace{-0.5cm}
\end{figure}

Scenarios involving blind spots are prevalent and pose significant challenges for autonomous driving. Ubiquitous occlusions, which can be categorized into static and dynamic occlusions (Fig. \ref{cover}a), limit the capabilities of perception and prediction and introduce high uncertainty.
 
There are several methods addressing occlusion uncertainty from different perspectives. Approaches based on reachable states analysis \cite{orzechowski2018tackling,set2021,risk2019ral,risk2023ral} achieve speed decision-making for static occlusions at intersections. However, ignoring active exploration ability, such methods lead to conservative and suboptimal behavior. When interacting with dynamic occlusions (Fig. \ref{cover}b) or with severely limited visibility, it is necessary to actively explore the conflict zone to reduce visibility uncertainty. 

To achieve active perception behavior, recent work as \cite{zhang2021safe} models it as a zero-sum game. However, such hand-crafted approaches face challenges in computational complexity and scenario scalability. Recent data-driven methods \cite{AVP,nvidia2023icra} also provide efficient solutions based on imitation learning and occluded agent trajectory prediction. Nevertheless, considering the high uncertainty and rarity of interactions with occluded agents in real scenarios, the cost of available data and expert bias severely constrain them.

Reinforcement Learning (RL) shows promise by automatically generating data through exploration and randomization. However, previous occlusion-aware RL research \cite{occlision_rl2018icra, occlision_rl2020iv, occlision_rl2023tits} mainly focused on speed decision-making at intersections under static occlusions. A general perception-aware learning solution for various scenarios with both dynamic and static occlusions faces challenges. 
Firstly, an efficient and lightweight representation of various occluded environments under dynamic and static occlusions is not well discussed. Secondly, using only speed control actions limits the agent's broader exploration, such as tentative probing. Under sparse rewards in occlusion-aware learning, fine-grained actions like acceleration and deceleration along the path time point \cite{occlision_rl2018icra}, or steering angles and pedals as in most existing RL works \cite{peng2021learning, zhang2021roach, chen2021interpretable, 10250993, gao2024enhance}, suffer from excessive variance in random exploration. 
Third, during exploring conflict zones from occlusion, RL agents may engage in extensive risky exploration. The lack of predictive ability may lead to low sampling efficiency and a lack of security guarantees.

% previous methods mainly discussed intersections under static occlusion. 
% Firstly, effectively representing blind spots is crucial for RL, but a general and efficient representation of various occluded environments under dynamic and static occlusions is not well discussed.
% As a self-reinforcing framework, Pad-AI is designed to be adaptable to both dynamic and static occlusions, diverse road scenarios, and unknown dynamics.
% which learns perception-aware decision-making through interacting with occluded agents. 

In this paper, we present Pad-AI (\textbf{P}erception-\textbf{A}ware \textbf{D}ecision-making from \textbf{A}gent \textbf{I}nteraction). The key insight is that, compared to previous occlusion-aware RL works, our Pad-AI achieves more efficient and general perception-aware exploration decision-making to handle high uncertainty from both dynamic and multiple static occlusions across various scenarios and tasks.
To address the occlusion uncertainty in various road structures, Pad-AI utilizes vectorized representation to efficiently characterize occluded environments. To provide efficient exploration for various occlusion scenarios, we propose semantic motion primitives (SMPs), which efficiently constrain RL agents to focus on high-level occlusion-aware decision-making, inspired by human perception-aware driving strategy. To avoid excessive risky exploration of conflict zones caused by various occlusion, prediction is integrated with RL based on a safe interaction mechanism, which further enhances the learning efficiency, risk-aware learning, and safety guarantee of RL. The major contributions are summarized as:

\begin{itemize} 
\item We propose an occlusion-aware RL method, in which learning efficiency is significantly improved through vectorized representation and semantic motion primitives (SMPs), which adapt to various driving scenarios under dynamic and static occlusion. 
% \item We propose a safety interaction mechanism that integrates RL and prediction to enhance sample efficiency by avoiding risky exploration and enabling risk-aware learning from reward. Moreover, it brings in security guarantees.
\item We propose a safety interaction mechanism that integrates RL and prediction to improve sample efficiency by avoiding risky exploration and enabling risk-aware learning, while also providing security guarantees.
\item We validated our method in challenging dynamic and static occlusion scenarios, achieving perception-aware expert-level performance to other strong baselines.
\end{itemize}

 \begin{figure*}[ht]
  \centering
  \includegraphics[width=0.8\textwidth,height=0.4\textwidth]{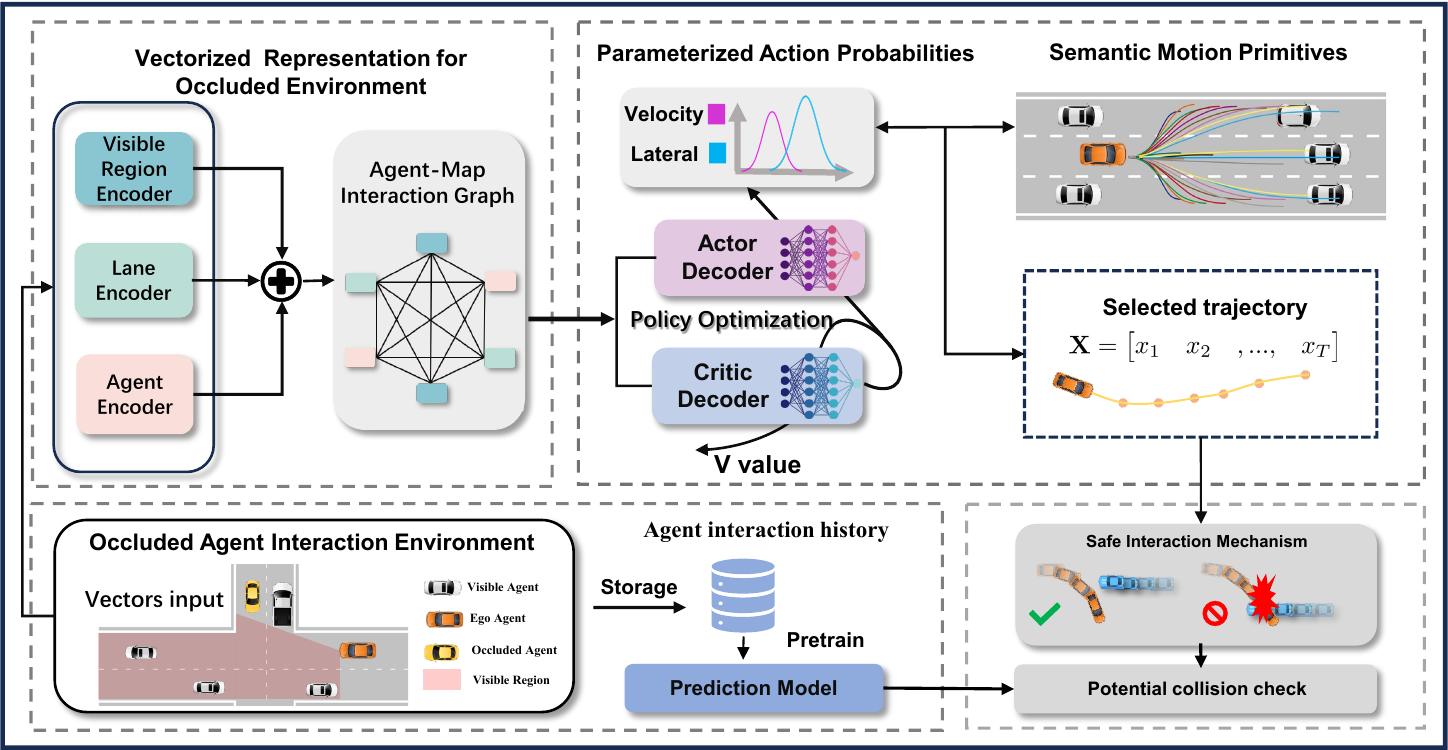}

  \caption{Schematic architecture: Vectorized observations from the occluded environment are encoded through a graph neural network. The actor decodes parameterized action probabilities, which are further mapped to the semantic motion primitives (SMPs). A safe interaction mechanism is engaged to avoid risky motion primitive exploration and security guarantees.}
  \label{Outlilne}
% \vspace{-0.15cm}
\vspace{-0.4cm}
\end{figure*}

\section{Related Work}

Prior work has proposed various decision-making methods addressing occluded agents from different perspectives. These differences are summarized in Table \ref{tab_I}.

A generic approach to deal with occlusions is to consider hypothetical objects in the occluded areas and adjust behavior based on their reachable states \cite{orzechowski2018tackling,set2021,risk2019ral,risk2020ICRA,risk2023ral,2023riskJournal}. The methods \cite{risk2019ral, risk2020ICRA, risk2023ral} use sampled potential hidden agent positions to approximate forward reachable states. Such works integrated the evaluated occlusion risks into the speed decision-making at intersections under static occlusions. \cite{set2021} uses an edge to classify static and dynamic phantom vehicles. The classification results were used to estimate the forward reachable sets (FRS) of all potential occluded objects. However, the reachability-based methods struggle to cover different types of occlusion situations. When interacting with dynamic occlusions or with severely limited visibility, it is necessary to actively explore the conflict zone to reduce visibility uncertainty.

% In contrast, we propose a data-driven approach to learn active perception decision-making under both dynamic and static occlusions from multi-agent interactions.

Additionally, previous works \cite{pomdp2018icra,pomdp2019iv,pomdp2022iv} formulate this problem as a Partially Observable Markov Decision Process (POMDP) with carefully designed observation models for occlusion in the belief state. Active perception behavior has been modeled in recent game theory work. \cite{zhang2021safe} models the interaction with hidden agents as a zero-sum game. However, these methods lack scenario scalability, as the environment gets more complex, they face challenges in computational complexity and conservative behavior.

More recently data-driven methods have been proposed. \cite{AVP} utilizes a flow-based generative model for imitation learning expert data. \cite{nvidia2023icra} predicts the possible trajectories of occluded agents based on datasets and integrates them into planning. However, due to the rarity of interactions with occluded agents in real scenarios, the scalability and performance of such methods are limited by the cost of available data and expert bias.

Reinforcement learning (RL) benefits from automatic data generation, offering a promising solution for occlusion-aware decision-making \cite{occlision_rl2018icra, occlision_rl2020iv, occlision_rl2023tits}. \cite{occlision_rl2018icra} used Deep-Q-Network (DQN) for occluded intersection crossing. \cite{occlision_rl2020iv} discussed the limitations of the grid representation \cite{occlision_rl2018icra} for various intersection topologies and proposed a conflict zone-based representation for occluded areas, but not adapted to intersections with dynamic occlusions. \cite{occlision_rl2023tits} employed a fully parameterized quantile network (FPQN) for reward estimation at occluded intersections. Previous works mainly focused on acceleration and deceleration actions under static occlusions. Steering angles and pedals offer a broad action space for RL agents in most recent RL works \cite{peng2021learning, zhang2021roach, chen2021interpretable, 10250993, gao2024enhance}. However, these fine-grained actions often suffer from excessive exploration variance under sparse rewards in occlusion-aware learning as we observed. Furthermore, the predictive ability has not been fully addressed to reduce risky exploration under perception uncertainty.
In this work, we explore efficiently vectorized representation, semantic motion primitives as actions, and integrating predictive capability into RL to provide an efficient and general solution across various scenarios under both dynamic and static occlusions.

\begin{table}[t]
  \centering
  \renewcommand{\arraystretch}{1.2}
  \newcommand{\mybox}[1]{\makebox[1.5em]{#1}}
  \scriptsize
  \setlength{\tabcolsep}{3pt}  % 控制列间距
  \rowcolors{2}{gray!10}{white}
  \begin{tabularx}{\columnwidth}{>{\centering\arraybackslash}p{0.7cm}>{\centering\arraybackslash}p{0.8cm}>{\centering\arraybackslash}p{1.0cm}>{\centering\arraybackslash}p{0.85cm}>{\centering\arraybackslash}p{0.7cm}>{\centering\arraybackslash}p{0.45cm}>{\centering\arraybackslash}p{0.8cm}>{\centering\arraybackslash}p{0.72cm}>{\centering\arraybackslash}p{0.7cm}}
    \toprule
     \raisebox{-0.7\height}{Method} & Learned \newline Model & Self \newline Supervised & Occluded \newline Obs. & Active \newline Explore &  \raisebox{-0.7\height}{Static} &  \raisebox{-0.7\height}{Dynamic} &  \raisebox{-0.7\height}{Intersec.} &  \raisebox{-0.7\height}{Overtake}\\
    \midrule
    \cite{orzechowski2018tackling}. & \mybox{\xmark} & \mybox{None} & \mybox{\cmark} & \textcolor{red}{\mybox{None}} & \mybox{\cmark} & \mybox{\xmark} & \mybox{\cmark} & \mybox{\xmark} \\
    
    \cite{risk2023ral} & \mybox{\xmark} & \mybox{None} & \mybox{\cmark} & \textcolor{red}{\mybox{None}} & \mybox{\cmark} & \mybox{\xmark} & \mybox{\cmark} & \mybox{\xmark} \\
    
    \cite{zhang2021safe} & \mybox{\xmark} & \mybox{None} & \mybox{\cmark} & \textcolor{customgreen}{\mybox{Active}} & \mybox{\cmark} & \mybox{\cmark} & \mybox{\cmark} & \mybox{\cmark} \\ 
    
    \cite{AVP} & \mybox{\cmark} & \mybox{\xmark} & \mybox{\cmark} & \textcolor{customgreen}{\mybox{Active}} & \mybox{\cmark} & \mybox{\cmark} & \mybox{\cmark} & \mybox{\cmark} \\
    
    \cite{occlision_rl2023tits} & \mybox{\cmark} & \textcolor{customgreen}{\mybox{Online}} & \mybox{\xmark} & \textcolor{customgreen}{\mybox{Active}} & \mybox{\cmark} & \mybox{\xmark} & \mybox{\cmark} & \mybox{\xmark} \\ 
    
    Ours & \mybox{\cmark} & \textcolor{customgreen}{\mybox{Online}} & \mybox{\cmark} & \textcolor{customgreen}{\mybox{Active}} & \mybox{\cmark} & \mybox{\cmark} & \mybox{\cmark} & \mybox{\cmark} \\
    \bottomrule
  \end{tabularx}
  \caption{Comparison of recent occlusion-aware planning methods for autonomous vehicles.}
  \vspace{-0.3cm}
  \label{tab_I}
\end{table}

\section{Method}

In this section, the occlusion-aware decision-making problem is formulated as an RL problem, represented as a Markov decision process. A Markov decision process can be defined as$\{\mathcal{S}, \mathcal{A}, \mathcal{T}, \mathcal{R}, \gamma \}$, a tuple consisting of states, actions, transition probabilities, rewards, and discount factor. Fig.~\ref{Outlilne} presents the total architecture of our method.

\subsection{Vectorized Representation of Occluded Environment}  \label{sec:obs}

\begin{figure}[t]
\centerline{\includegraphics[width=0.48\textwidth,height=0.25\textwidth]{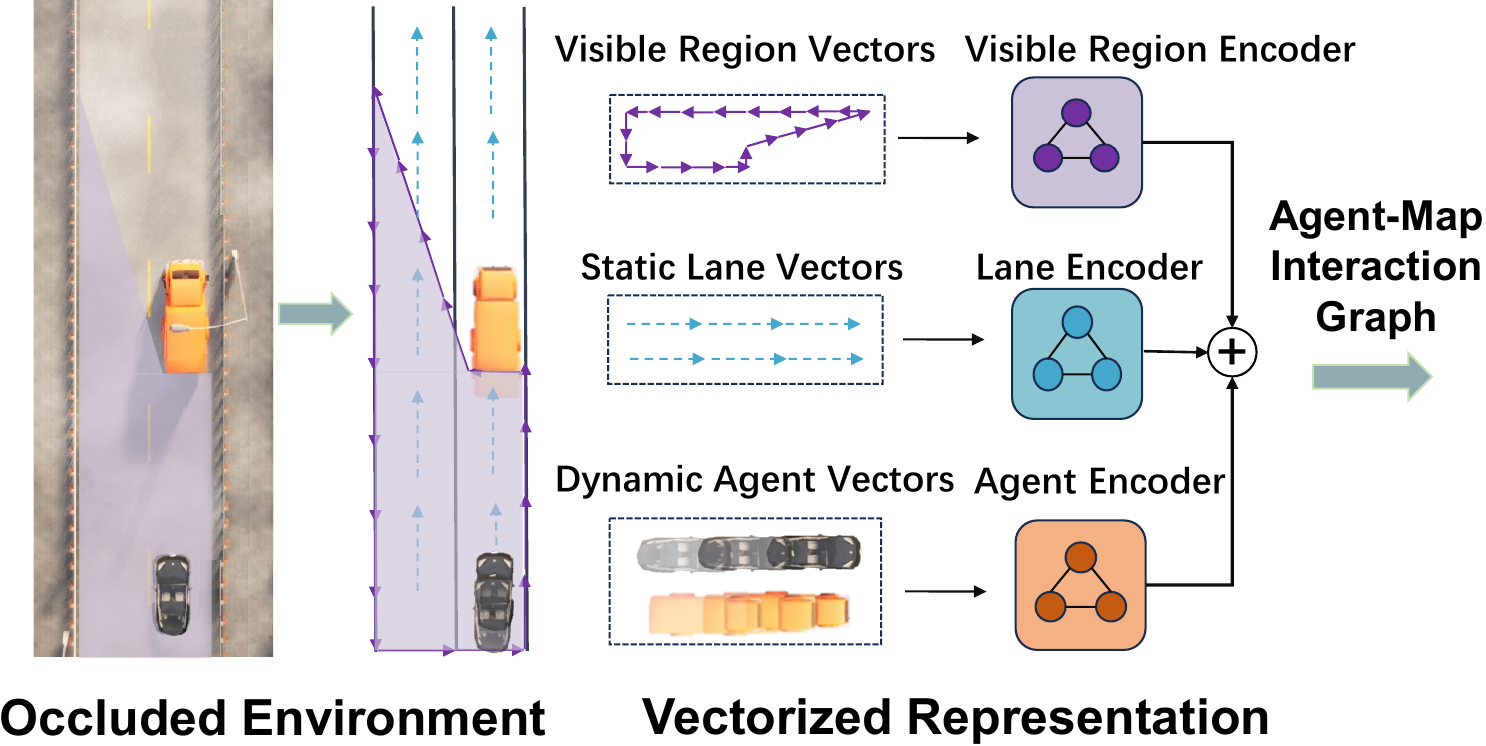}}
% \centerline{\includegraphicss[width=0.485\textwidth,height=0.4\textwidth]{method_obs.jpg}}
\caption{Vectorized representation of occluded environments, including visible region, lane lines, and agent history.} % Add a caption to the figure
\label{obs}
\vspace{-0.5cm}
\end{figure}

State representation is crucial for RL. To better represent blind spots in RL, we adopt vectorization to represent the occluded environment. Our vectorized observation, as depicted in Fig. 3, includes three sequences of vectors: visible region (VR), static lane lines, and dynamic agent history. Unlike ignoring visible region as input \cite{occlision_rl2023tits}, our vectorized representation explicitly represents the ego agent’s blind spots in sequences of vectors,ensuring a lightweight network. Additionally, our vectorization blind spots representation enables adaptability to various road structures under interaction with both dynamic and static occlusion.

Specifically, as shown in Fig. \ref{obs}, the visible region of the ego agent can be calculated through ray tracing, as done in previous occlusion-aware works \cite{zhang2021safe,occlision_rl2018icra}. We approximate the VR as a polygonal region $\mathcal{P}$ and represent the polygonal region by its boundary. Given a starting point and direction, this boundary can be a vectorized form of a closed polyline $\mathbf{p} = [p_0, p_1, ..., p_n]$, where $p_{i}$ is the vertice of the polygon. This polyline representation forms a graph structure, where each vector $p_{i}$ belonging to the polyline $\mathbf{p}$ serves as a node in the polyline graph. Each node's features are defined as $p_{i} = (p_i^s, p_i^e)$ (for $i=0,1,2,...,n$), where $p_i^s$ and $p_i^e$  represent the starting and ending coordinates of the vector respectively.

Similarly, in each observation step, for lane lines within the visible range, we uniformly sample key points from splines at the same spatial distance and connect adjacent key points in sequence to form vectors $\mathbf{l} = [l_0, l_1, ..., l_n]$. Each node feature in the lane graph is defined as $l_{j} = (l_j^s, l_j^e)$ ( $\forall j=0,1,2,...,n$), where $l_i^s$ and $l_i^e$ represent the starting and ending coordinates of the lane vector. For observed agent trajectories, key points are sampled at a fixed time interval (0.1 seconds) starting from t = 0, and they are connected into vectors $\mathbf{h} = [h_0, h_1, ..., h_n]$. Each node feature in the agent graph is defined as $h_{k} = (x_k, y_k, v_k, \psi_k, t)$ ($\forall k=0,1,2,...,n$), which represents the agent's position coordinates, velocity, yaw angle, and timestamps respectively.

After being vectorized into graph structures, each semantic element is encoded through a hierarchical graph neural network \cite{gao2020vectornet}. Firstly, the local features of each element are encoded by a subgraph network and max-pooling operation to aggregate the local features. Next, these local features serve as nodes in the global fully connected Graph Attention Network (GAT). The final features are obtained through the self-attention operation.

\subsection{Semantic Motion Primitives} \label{sec:smp}

To provide general and efficient active exploration for various occlusion scenarios and tasks, we propose semantic motion primitives (SMPs) in Fig. \ref{Outlilne}. Due to the significant uncertainty from interactions with dynamic and static occlusions, the fine-grained single-step actions lead to excessive variance in random exploration. However, each of our primitives represents a semantic intention short-term trajectory, such as tentative probing or yielding, enabling the agent to focus on high-level active perception exploration.

The action probability distribution from the RL directly maps to the control space of primitives.The generation of primitives is independent of tasks and scenarios, avoiding the complex motion primitives design. Specifically, we represent vehicle trajectories along the longitudinal and lateral axes using local coordinates of the reference road origin and path. Motion primitives are generated by sampling target waypoints and control-optimal polynomial trajectories. Target waypoints are defined by $\Phi_{\mathrm{e}}= \{v_{x\mathrm{e}},a_{x\mathrm{e}},d_{y\mathrm{e}},v_{y\mathrm{e}},a_{y\mathrm{e}}\}$, where $x$ represents the longitudinal component, $y$ represents the lateral component, and $d$, $v$, $a$ respectively denote the displacement, velocity, and acceleration attributes of the waypoint. Diverse polynomial trajectories are generated by sampling target states from the target space $\Phi_{\mathrm{e}}$. Boundary conditions for the primitive are given by the starting waypoint $\Phi_{\mathrm{s}} = \{d_{x\mathrm{s}},v_{x\mathrm{s}},a_{x\mathrm{s}},d_{y\mathrm{s}},v_{y\mathrm{s}},a_{y\mathrm{s}}\}$ and the target waypoint $\Phi_{\mathrm{e}}$, where $\mathrm{s}$ denotes the starting position.

We set the duration of a primitive to $\mathrm{\textit{T}} =3~\mathrm{s}$, with an interpolation interval  $\Delta \tau = 0.1~\mathrm{s}$. The lateral and longitudinal movements are described by Eq.~\ref{eq:polyfit}. Interestingly, the polynomial interpolation is proven to be control-optimal and we refer interested readers to \cite{werling2010optimal} for details.
% mathematical 
\begin{equation}    
\left\{    
\begin{aligned}        
\mathbf{x}(\tau) &= a_0 + a_1\tau + a_2\tau^2 + a_3\tau^3 + a_4\tau^4 \\        \mathbf{y}(\tau) &= b_0 + b_1\tau + b_2\tau^2 + b_3\tau^3 + b_4\tau^4 + b_5\tau^5,   
\end{aligned}    \right.\label{eq:polyfit}
\end{equation}
% $\tau \le \mathrm{T}$
where $\tau \le \mathrm{\textit{T}}$ represents the time step of the driving primitive, $a_i~(i=0,1,2,3,4)$ and $b_j~(j=0,1,2,3,4,5)$ are the polynomial coefficients solved from the boundary conditions.  Since the generated primitives should be smooth and dynamically feasible, continuity in path velocity, acceleration is required. Fig. \ref{Outlilne} illustrates the trajectory generation process, where only $v_{x\mathrm{e}}$ and $d_{y\mathrm{e}}$ vary, while others are held at zero.

\subsection{Safety Interaction Mechanism}
\label{sec:sim}

\begin{figure}[t]
\centerline{\includegraphics[width=0.48\textwidth,height=0.17\textwidth]{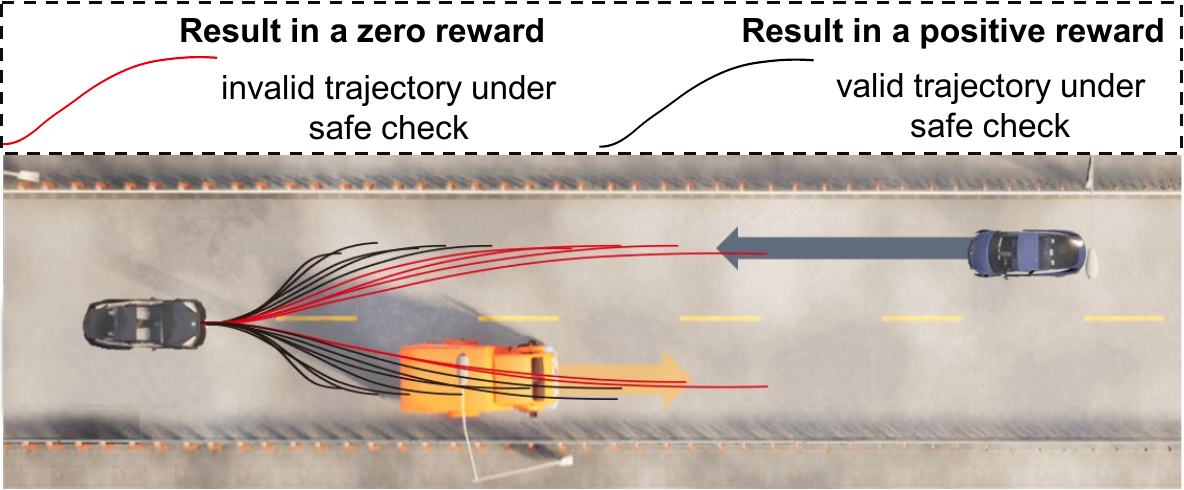}}
\caption{Safety Interaction Mechanism} % Add a caption to the figure
\label{sim}
\vspace{-0.7cm}
\end{figure}

During the active exploration with occlusion, traditional RL faces excessive risky exploration of conflict zones. Therefore, prediction is integrated with RL by the mechanism (Fig. \ref{sim}) to avoid risky exploration and improve sampling efficiency in RL, which further enhances risk-aware learning on the reward feedback and provides security guarantees.

Specifically, our mechanism is conducted on the planned trajectory of the ego vehicle and the predicted trajectories of other vehicles. We use a neural network to predict other vehicles. Within one planning cycle, to check the safety of each explored primitive, the predicted trajectories of other vehicles are continuously updated. During the execution of the selected primitive, it will be terminated if it no longer satisfies the constraints of the security boundary calculated from Responsibility-Sensitive Safety (RSS) \cite{shalev2017formal} concerning other vehicles' short-term predicted trajectories. We use loose RSS parameters and short-term predictions to make sure that only extremely dangerous actions are penalized. Subsequently, while avoiding excessive risky exploration, the primitive identified as risky will receive zero rewards for risk-aware learning. In contrast, the fully executed primitive will receive positive rewards to encourage the agent.

\subsection{Reinforce through Agent Interaction}
In this section, we introduce the method of learning perception-aware behavior through interacting with potentially occluded agents.
To collect observations to train our learned model, we use predefined policies for non-ego
agents, which are designed to generate all plausible modes of realistic driver behavior in the scenario.

\textbf{Agent Interaction:} Interactive agents are randomly spawned in occluded regions. During each episode, the occluded agent may rush into the AV's field of view at diverse times and speeds, sampled to cover all plausible modes of realistic driver behavior. To speed up occlusion-aware learning, areas near the edge of the ego vehicle's blind spot are given higher probabilities during early training.

\textbf{Reinforcement Learning:} As Fig. \ref{Outlilne}, the process begins with obtaining a vectorized representation and feature encoding of the current state (Section \ref{sec:obs}). The encoded features are input into the actor decoder to generate parameterized Gaussian action probabilities, $\pi(a|s)$. These probabilities are mapped and normalized to the end conditions of semantic motion primitives (SMPs) (Section \ref{sec:smp}). The primitive with the maximum probability is executed for a horizon length of $T_{exec}<T$. During each interval of $T_{exec}$, a safety interaction mechanism (Section \ref{sec:sim}) is engaged to assess risk check. While training, steps terminated in collisions receive negative rewards, risky steps receive zero rewards, and fully executed primitives receive positive rewards. The critic outputs the state value. Based on the risk-aware rewards, the actor and critic modules are alternately optimized by Proximal Policy Optimization (PPO) \cite{schulman2017proximal}. Our observations show that SMPs and the safety interaction mechanism greatly enhance RL policy sample efficiency and risk awareness learning. Details of the ablative study are in Section~\ref{sec:exp}.

%, an RL algorithm based on the actor-critic framework. PPO achieves a balance between exploration and exploitation by clipping the policy update, preventing large policy shifts that could lead to performance degradation. This approach ensures a more stable and reliable training experience.

% 先不强调
% Each SMP is depicted as a trajectory spanning a horizon of \mathrm{T} time steps.

\textbf{Prediction Model Pretraining:} In the training process of Pad-AI, the safety interaction mechanism introduced in Section \ref{sec:sim} relies on inputs from the agent's motion predictions, provided by the pre-trained prediction model. We utilize historical motion observations collected from the interaction between RL and multiple agents as the dataset for the prediction model. This strategy integrates prediction with RL and renders a fully auto-labeling process for prediction tasks. The prediction model can be fine-tuned during the RL training. To streamline the design, the prediction model adopts the same observation and encoder design as RL.

% \vspace{-0.08cm}
\section{Experiments}\label{sec:exp}
\subsection{Simulation Platform and Environment}

Inspired by previous occlusion-aware work and real-world scenarios, we evaluate Pad-AI's performance in challenging interaction scenarios under dynamic and severe static occlusion in the CARLA simulator \cite{Dosovitskiy17}. All the evaluation experiments are implemented on a desktop with an Intel i7-12700 CPU and an Nvidia GeForce RTX 3060 Ti GPU.

\textbf{Two-lane overtake:} Ego vehicle is obstructed by a slow-moving truck on an undivided highway. As a large dynamic occlusion, the truck not only blocks the visibility in both the current and opposing lanes.

\textbf{Occluded T-Intersection:} Ego agent navigates the T-intersection, where oncoming traffic is occluded by severe occlusion, potentially hiding agents in the right blind spot.

\textbf{Occluded Crossroad:} Oncoming traffic is blocked by multiple severe static occlusions in a crossroad, with one or more agents potentially emerging from either side.

\subsection{Implementation Details}
Based on RLLib \cite{liang2018rllib} (a distributed learning system that allows large-scale parallel experiments), parallel interactive experiments with different random seeds can be performed on multiple environments to speed up the interaction process. During the learning process, each episode will terminate either when the autonomous vehicle reaches its destination, collides, times out, or runs off the roadway.

\textbf{Metrics:} We evaluate our method over 100 episodes with a balanced split of modes for each scenario, comparing it to existing strong baselines using the following metrics.

\textit{Success rate:} The percentage of episodes in which the AV successfully reaches the goal without any collisions. A higher success rate refers to better policy performance.
\textit{Collision rate:} The percentage of collision episodes in the whole testing process. A lower collision rate indicates higher safety.
\textit{Speed:} The average speed during successful episodes. A higher speed indicates better time efficiency.

\subsection{Baselines}

EM Planner (EMP) \cite{fan2018baidu}: A widely used planning method in industry that observes the behavior of visible agents.
% However, this method ignores the uncertainty from the occlusion blind area, leading to overconfident behavior under high perception uncertainty.

Reachable State Analysis (RSA) \cite{set2021}: A SOTA occlusion-aware planning method based on reachable risk assessment.

% over-approximates the state of hidden agents by reachable set analysis for safe decision-making.
% ignoring active exploration actions to reduce perception uncertainty, leading to conservative suboptimal behavior.

Game-theoretic (SOAP) \cite{zhang2021safe}: One SOTA occlusion-aware decision-making method (SOAP) generates perception-aware expert driving strategies based on zero-sum game theory. 

\begin{figure}[t]
% \centerline
% {\includegraphics[width=0.48\textwidth]{FRS_overtake_v4.pdf}}
{\includegraphics[width=0.48\textwidth]{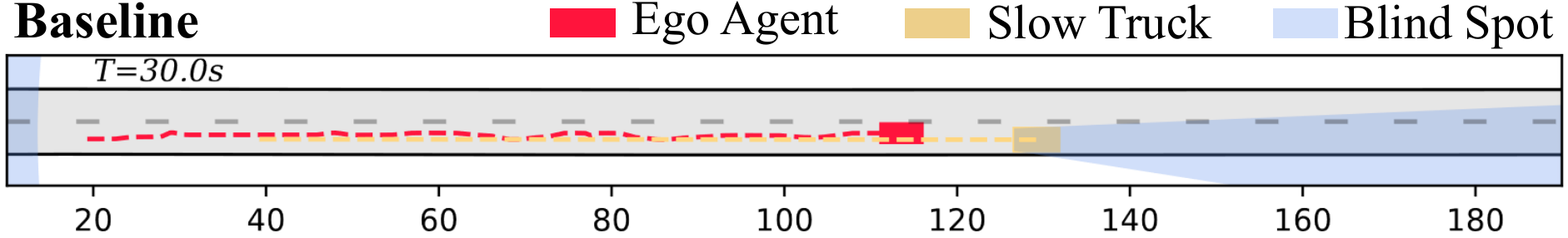}}
\caption{RSA driver continuously follows the slow truck, limiting the ability to change lanes actively.} % Add a caption to the figure
\label{FRS}
\vspace{-0.5cm}
\end{figure}

\begin{figure}[t]
     % \centering
    % \vspace{-0.1cm}
    \begin{minipage}[b]{0.48\textwidth}
        % \centering
        % \includegraphics[width=1.0\textwidth]{overtake_a_v3.pdf}
        \includegraphics[width=1.0\textwidth]{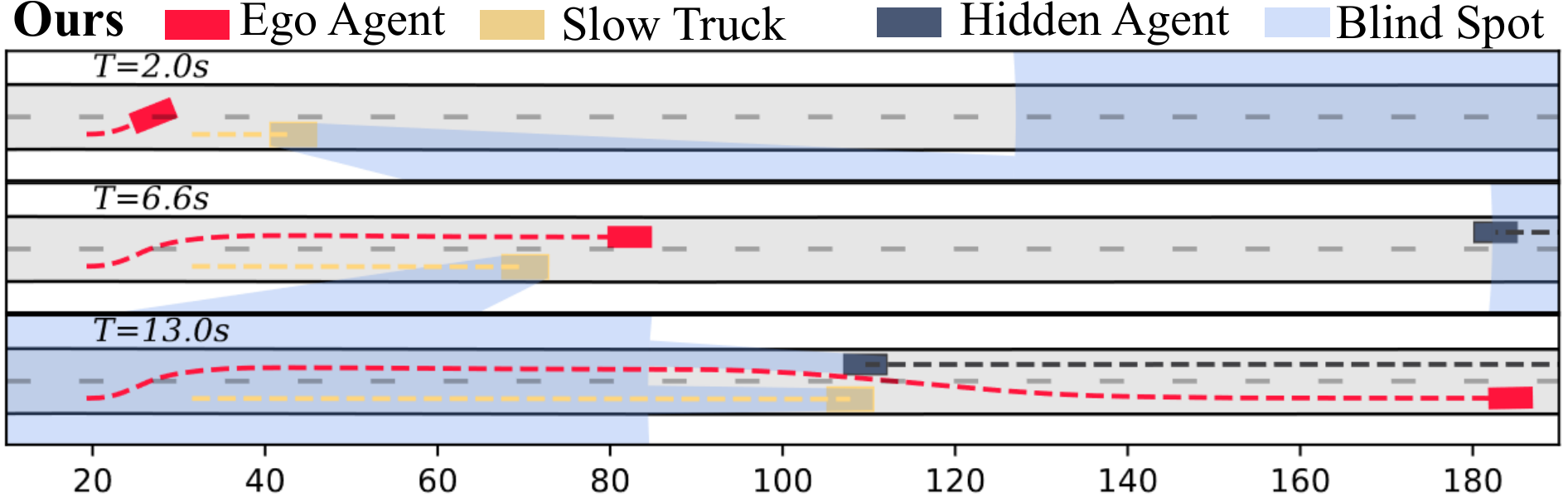}
        (a) Actively probe, swiftly overtake based on the updated observations, and merge back to avoid the oncoming agent.
        \vspace{-0.1cm}
        \label{overtake_a}
    \end{minipage}
    % \hfill
    \begin{minipage}[b]{0.48\textwidth}
        % \centering
        % \includegraphics[width=1.0\textwidth]{overtake_b_v3.pdf}
        \includegraphics[width=1.0\textwidth]{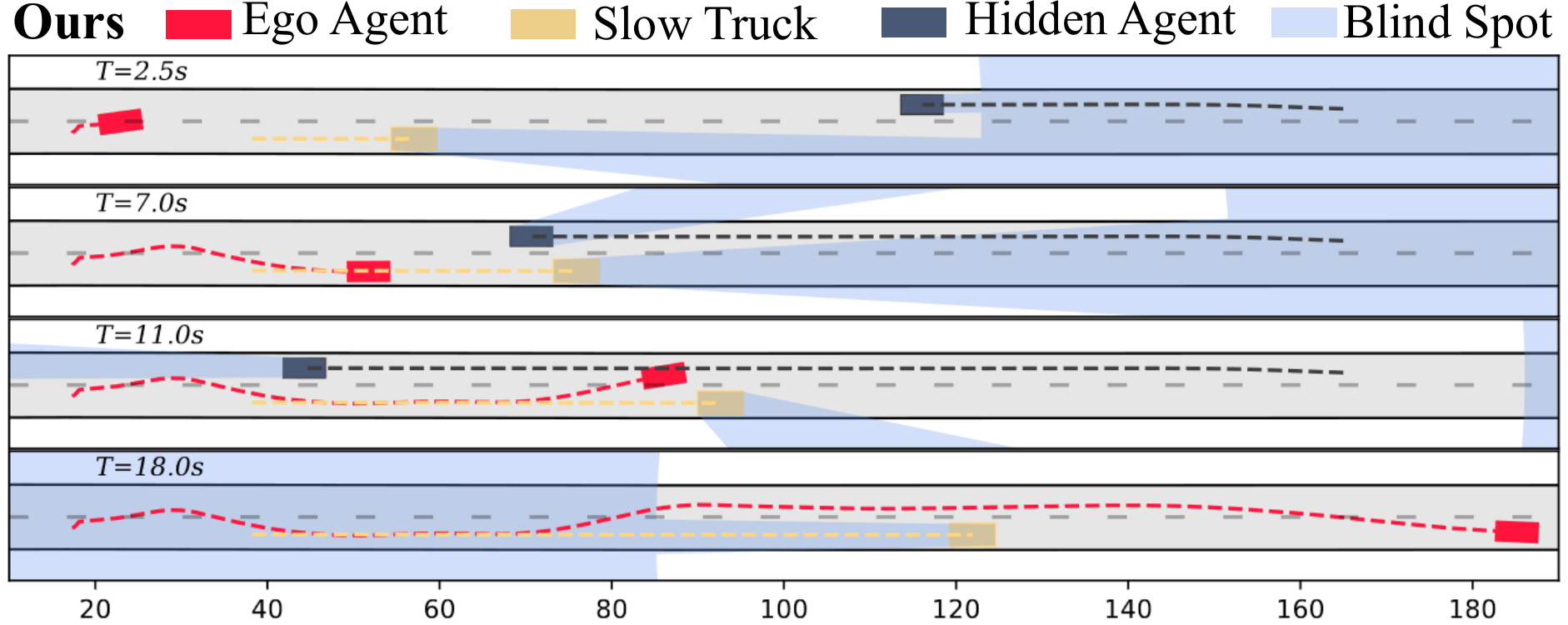}
        (b) Actively probe, abort the overtaking after observing the oncoming agent, and make the next attempt.
         % \vspace{-0.1cm}
        \label{overtake_b}
    \end{minipage}
    \caption{Our vehicle actively probes to reduce uncertainty. Further decisions are based on updated observations: (a) swiftly overtake and avoid being trapped parallel to the truck; (b) abort the overtaking by the safe interaction mechanism.}
    \label{overtake_RL}
    \vspace{-0.5cm}
\end{figure}

\begin{table*}[ht]
% \centering
% \setlength{\tabcolsep}{2.5pt} % 默认是6pt，这里减少了列之间的间距
% \renewcommand{\arraystretch}{1.1} % 默认是1，这里稍微增加了行高
\begin{tabular}{@{}lcccccccccccc@{}}
\toprule
& \multicolumn{3}{c}{Two-lane overtake} & \multicolumn{3}{c}{T-Intersection } & \multicolumn{3}{c}{Crossroad} \\
\cmidrule(lr){2-4} \cmidrule(lr){5-7} \cmidrule(lr){8-10}
Method & SR(\%)$\uparrow$ & CR(\%)$\downarrow$ & Speed(m/s)$\uparrow$ & SR(\%)$\uparrow$ & CR(\%)$\downarrow$ & Speed(m/s)$\uparrow$ & SR(\%)$\uparrow$ & CR(\%)$\downarrow$ & Speed(m/s)$\uparrow$ \\
\midrule
EM Planner (EMP) & 63 & 37 & 12.3 & 54 & 46 & 12.1 & 23 & 77 & 11.7 \\ 
Reachable State Analysis (RSA) & 100 & 0 & 5.9 & 23 & 0 & 5.9 & 12 & 0 & 5.1 \\ 
Game-theoretic (SOAP)& 100 & 0 & 9.4 & 100 & 0 & 8.8 & 67 & 33 & 10.5 \\ 
Pad-AI (ours)* & 100 & 0 & 9.5 & 100 & 0 & 9.3 & 100 & 0 & 7.1 \\ 
\bottomrule
\end{tabular}
\caption{Compare our Pad-AI with baselines in a closed-loop evaluation.}
\label{tab:exp_data}
\vspace{-0.3cm}
\end{table*}

\begin{figure*}[htpb]
% \centerline{\includegraphics[width=0.98\textwidth,height=0.15\textwidth]{T_v9.pdf}}
% \centerline{\includegraphics[width=0.8\textwidth,height=0.13\textwidth]{T_v9.pdf}}
\centerline{\includegraphics[width=0.8\textwidth,height=0.13\textwidth]{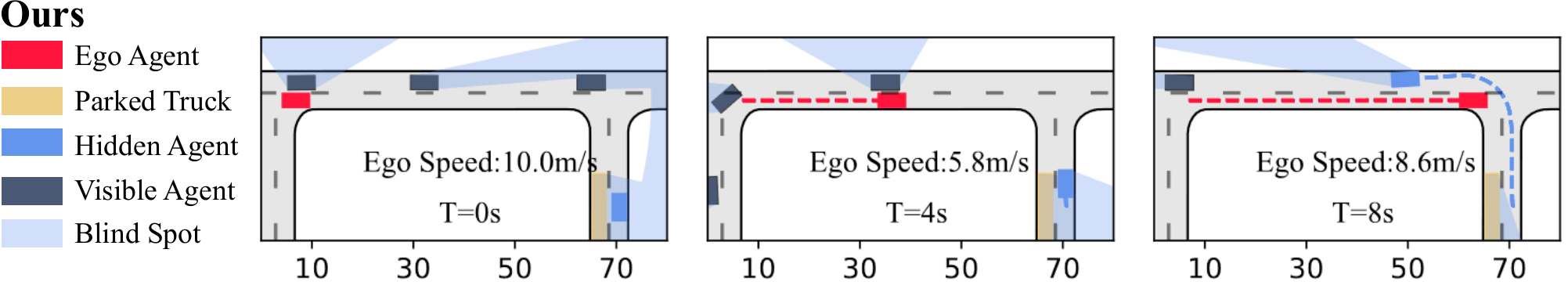}}
\hspace{0.8cm}(a) Approaching the occluded T-intersection \hspace{0.1cm} (b) Occluded agent emerges \hspace{0.1cm} (c) Clear the T-intersection after yielding
\caption{Approaching the blind T-intersection (a), our vehicle actively slows down in advance to reduce the risk of occlusion and responds to emerging hidden traffic (b). As the blind spot narrows, our vehicle accelerates to pass the intersection (c).}
\label{T}
\end{figure*}

\begin{figure*}[htpb]
% \centerline{\includegraphics[width=0.85\textwidth,height=0.20\textwidth]{cross_v9.pdf}}
\centerline{\includegraphics[width=0.85\textwidth,height=0.20\textwidth]{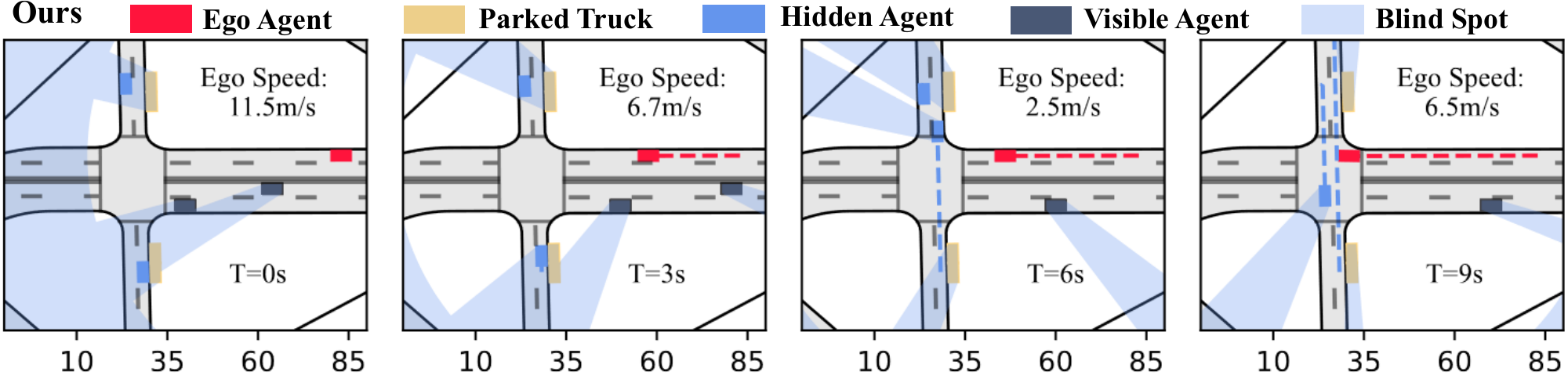}}
% \hspace{0.8cm}
(a) Approaching a crossroad 
(b) Agent emerges from left
(c) Agent emerges from right
(d) Clear the crossroad after yielding
\caption{Approaching a blind crossroad (a), our vehicle actively slows down in advance for further potential interaction with multiple occluded agents (b). The agents emerge successively from the left and right, and our vehicle further decelerates and yields them successively to ensure safety (c). As the blind spot narrows, our vehicle accelerates to pass (d).}
\label{cross}
% \centerline{\includegraphics[scale=0.5,trim=0 0 0 0]{cross_v2.pdf}}
\vspace{-0.4cm}
\end{figure*}

\subsection{Results and Analysis}

\textit{Qualitative analysis:}
The detailed experiments are in the attached video, and the quantitative results are in \ref{tab:exp_data}. In our evaluation, ignoring occlusion uncertainty, EMP often gets trapped by oncoming traffic, and the truck during overtaking, passes severely occluded intersections too quickly, all of which increases collision risks. Based on the estimated reachable set, RSA exhibits conservative and suboptimal behavior, hindering active exploration and causing actions like avoiding overtaking slow trucks (as shown in Fig. \ref{FRS}) and crossing very slowly or even freezing under severe occlusion.

In contrast, Pad-AI generates perception-aware expert-level behaviors. As Fig. \ref{overtake_RL}, when occluded by a slow truck, Pad-AI actively slows down to probe and reduce perception uncertainty. Decisions are then made based on the updated field of view. If no oncoming traffic is observed (Fig. \ref{overtake_RL}a), the agent swiftly overtakes and merges back to avoid being trapped parallel to the truck. Even if oncoming traffic appears during probing (Fig. \ref{overtake_RL}b), the safe interaction mechanism aborts overtaking in advance, and the agent makes the next attempt. Approaching the severely occluded T-intersection (Fig. \ref{T}), our agent actively slows down in advance to reduce occlusion risks, further decelerating if a hidden agent emerges. As visibility improves, it gradually accelerates through the intersection. Pad-AI's active uncertainty reduction is key for safe and optimal decision-making.

Our Pad-AI also scales to scenarios with multiple occluded agents under limited visibility. In Fig. \ref{cross}, it slows down in advance, enabling potential interaction with multiple occluded agents, similar to T-intersections.
Even when hidden agents emerge from both sides, our vehicle decelerates and yields sequentially to ensure safety.

\textit{Quantitative analysis:} Table \ref{tab:exp_data} shows that, unlike EMP, which ignores occluded uncertainty, Pad-AI reduces collision rates by 37\%, 46\%, and 77\%. Compared to RSA, Pad-AI improves speed efficiency by 61\%, 52\%, and 37\% and significantly improves the pass rate under severe occlusion. While SOAP performs similarly to Pad-AI in truck overtaking and T-intersections, it exhibits a reduced pass rate when extended to multiple occluded agents at crossroads. Due to game modeling limitations. SOAP leads to overconfident behavior. Our Pad-AI, as a learning-based method, performs well when extended to interaction with multiple severe occlusions.

More importantly, SOAP also incurs significant computational costs and setup requirements for each scenario. Its search-based solving method 
leads to intensive trajectory computation costs and map discretization for each scenario.  Based on SOAP's open-source code, the average computation times for planned trajectories are 328.14s, 8.76s, and 18.92s for overtaking, T-intersections, and crossroads, respectively, making real-time application difficult. In contrast, our Pad-AI computes in 0.046s, 0.048s, and 0.091s for these scenarios, enabling real-time applications. Moreover, Pad-AI’s lightweight vectorized observation is easily adaptable to various environments with minimal overhead.

\subsection{Ablation Study}

\begin{figure}[t]
% \centerline
{\includegraphics[width=0.48\textwidth]{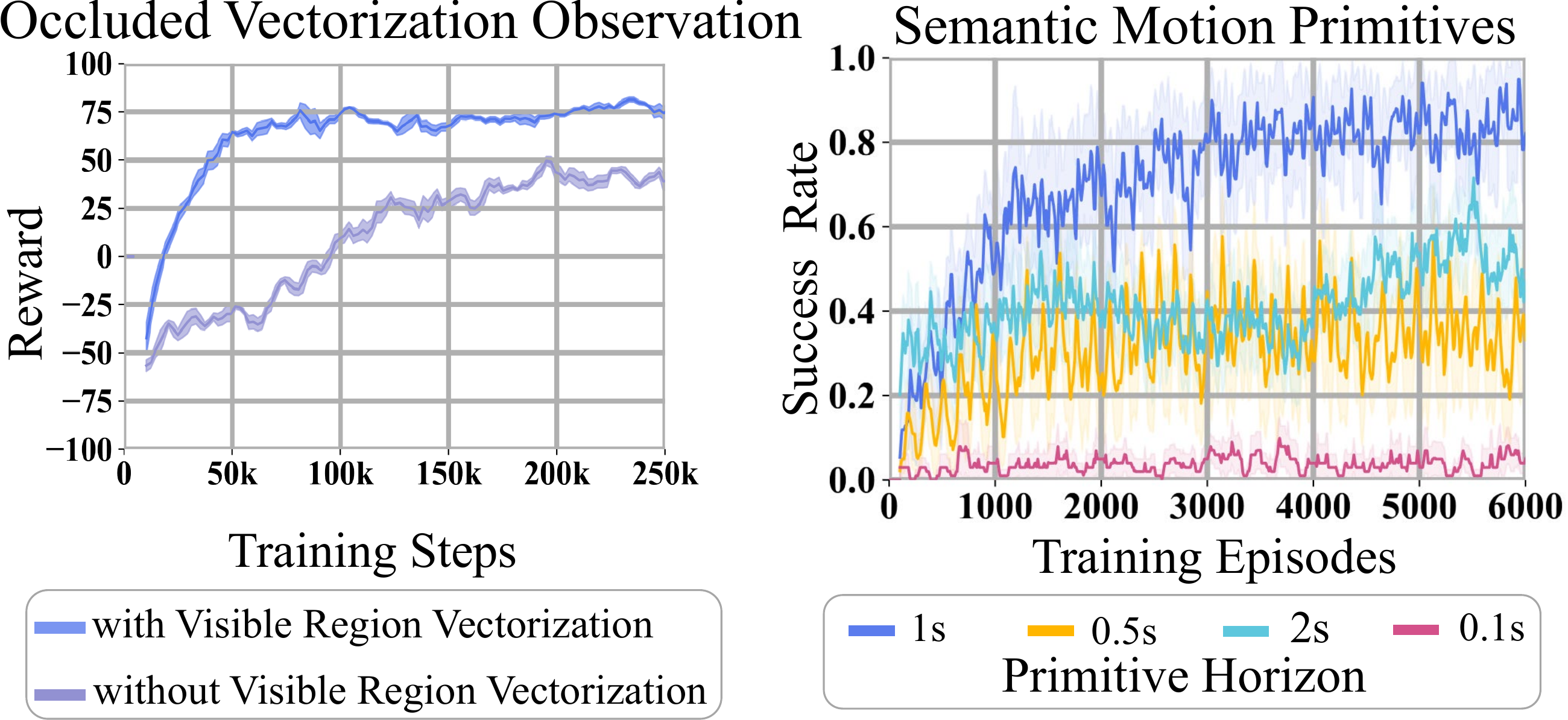}}
\caption{Ablation analysis the influence of occluded vectorization observation (left) and semantic motion primitives (right).} % Add a caption to the figure
\label{ICRA_ablate}
\vspace{-0.6cm}
\end{figure}

\textit{Influence of occluded vectorization observation:}  
We ablate the visible region input in overtaking scenarios, evaluating with the reward (Fig. \ref{ICRA_ablate}, left) during training, and the test success rates which dropped by 42\%. Additionally, replacing the discrete grids representation method applied in crossroads from \cite{occlision_rl2018icra} led to a 31\% decrease in test success rate on occluded crossroad scenarios. These results show that our occluded representation, adapted to various road structures, effectively represents occluded environments.

\textit{Influence of the semantic motion primitive:} We ablate the length of primitive execution parameter $T_{exec}$ in overtaking scenarios by success rate during training. As in Fig. \ref{ICRA_ablate} (right), $T_{exec}$ = 1s provides a good balance. With $T_{exec}$ = 0.1s, Pad-AI outputs fine-grained control actions, leading to poor learning efficiency. For $T_{exec}$ = 0.5s or 1s, we see continuous improvement in performance, highlighting the benefits of semantic long-term actions. However, a longer horizon ($T_{exec}$ = 2s) makes the agent less responsive to accidents and emergencies, reducing performance.

\textit{Influence of safety interaction mechanism:} We ablate the safety interaction mechanism on overtaking scenario with the test success rate decrease of 23\%, indicating its effectiveness in risk-aware learning and ensuring timely responsiveness to emergencies arising from perception uncertainty.

% \vspace{-0.1cm}
\section{CONCLUSIONS}
In this paper, we introduce Pad-AI, an efficient RL framework for learning perception-aware decision-making for various scenarios under dynamic and static occlusions. We propose the vectorized representation for the occluded state and SMP for action, which both enhance learning efficiency at a large margin. We further propose a safe interaction mechanism that integrates prediction with RL and avoids risky yet expensive exploration. We demonstrate the expert-level performance of Pad-AI compared to other strong baselines under dynamic and static occlusions. In the future, we will conduct real-world tests for the Pad-AI framework.

\bibliographystyle{plain}
\bibliography{ref}

\end{document}